\documentclass[runningheads]{llncs}
\usepackage[T1]{fontenc}
\usepackage{graphicx,verbatim}
\usepackage{multirow}
\usepackage{booktabs}
\usepackage{makecell}
\begin{document}
\title{Clinical-Prior Guided Multi-Modal Learning with Latent Attention Pooling for Gait-Based Scoliosis Screening}
%

\author{
Dong Chen\inst{1, 2, 3, 4} \and
Zizhuang Wei\inst{4} \and
Jialei Xu\inst{4} \and
Xinyang Sun\inst{4} \and
Zonglin He\inst{3} \and
Meiru An\inst{3} \and
Huili Peng\inst{1, 3} \and
Yong Hu\inst{1, 3} \and
Kenneth MC Cheung\inst{1, 2, 3}
\institute{
Orthopaedic Centre, The University of Hong Kong - Shenzhen Hospital, Shenzhen, China \and
Translational Medicine Centre, The University of Hong Kong - Shenzhen Hospital, China \and
Department of Orthopaedics and Traumatology, Li Ka Shing Faculty of Medicine, The University of Hong Kong, Hong Kong, China \and
Huawei, China
}
\email{olichen@connect.hku.hk; cheungmc@hku.hk}
}
  
\maketitle              
\begin{abstract}
Adolescent Idiopathic Scoliosis (AIS) is a prevalent spinal deformity whose progression can be mitigated through early detection. Conventional screening methods are often subjective, difficult to scale, and reliant on specialized clinical expertise. Video-based gait analysis offers a promising alternative, but current datasets and methods frequently suffer from data leakage, where performance is inflated by repeated clips from the same individual, or employ oversimplified models that lack clinical interpretability. To address these limitations, we introduce ScoliGait, a new benchmark dataset comprising 1,572 gait video clips for training and 300 fully independent clips for testing. Each clip is annotated with radiographic Cobb angles and descriptive text based on clinical kinematic priors. We propose a multi-modal framework that integrates a clinical-prior-guided kinematic knowledge map for interpretable feature representation, alongside a latent attention pooling mechanism to fuse video, text, and knowledge map modalities. Our method establishes a new state-of-the-art, demonstrating a significant performance gap on a realistic, non-repeating subject benchmark. Our approach establishes a new state of the art, showing a significant performance gain on a realistic, subject-independent benchmark. This work provides a robust, interpretable, and clinically grounded foundation for scalable, non-invasive AIS assessment. Code will be released upon publication.


\end{abstract}
\section{Introduction}

Adolescent idiopathic scoliosis (AIS) is a structural, lateral curvature of the spine with vertebral rotation, affecting approximately 5\% of children globally \cite{Cheng2015,Jin2018Screening}. If untreated, it can result in significant health consequences including chronic back pain and psychosocial distress \cite{Altaf2013Adolescent}. While diagnostic gold standard, defined by Scoliosis Research Society, relies on radiographic measurment of a coronal Cobb Angle (CA) exceeding 10 degrees \cite{Negrini2016SOSORT}. The necessity of frequent monitoring raises concerns regarding cumulative radiation exposure, underscoring the need for non-invasive and scalable screening alternatives. Current standard screening methods, such as the Adam's Forward Bending Test combined with scoliometer, are commonly employed in community settings. However, their effectiveness is inconsistent and influenced by subjective interpretation and factors like obesity \cite{Luk2010Clinical,Coelho2013Scoliometer,Margalit2017Body}. Furthermore, these methods pose practical challenges for large-scale implementation due to their dependence on clinical expertise, cost, and privacy concerns \cite{Luk2010Clinical}.

\begin{figure*}[ht]
    \includegraphics[width=\textwidth]{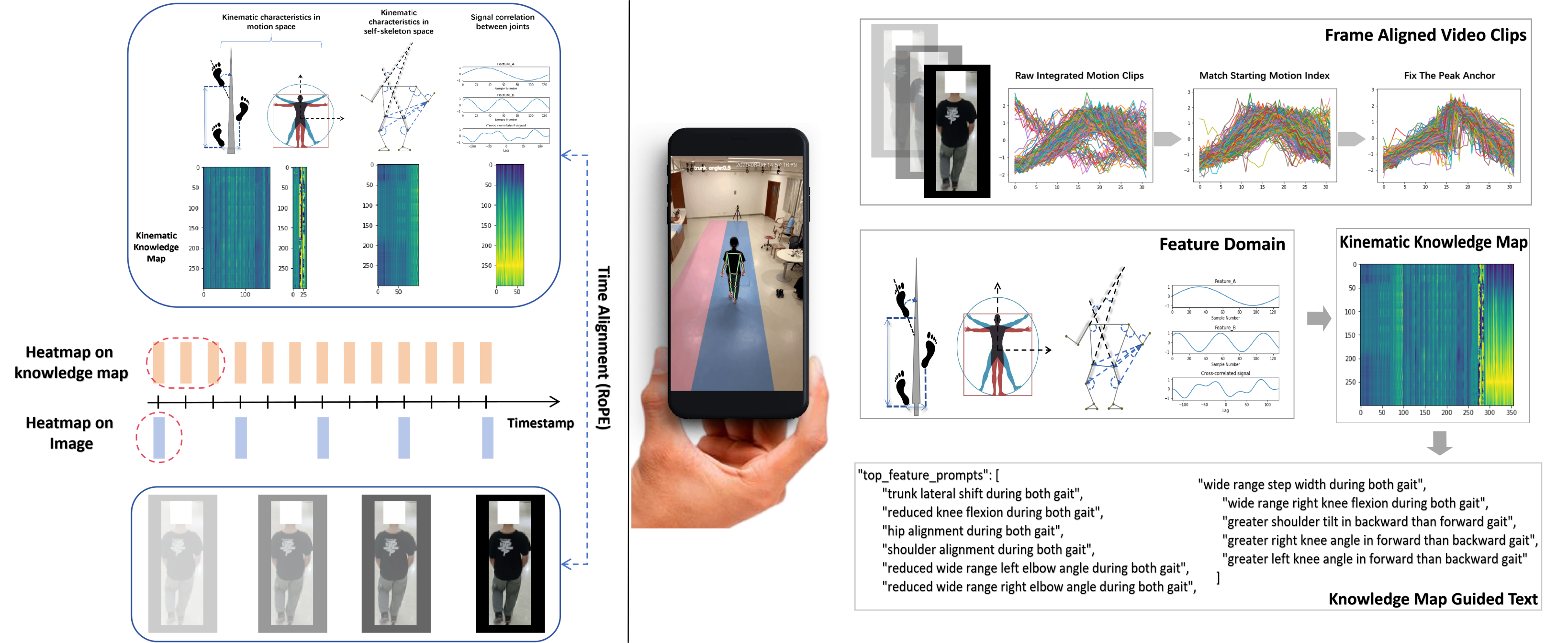}
    \caption{ScoliGait system for multi-modal gait analysis from mobile video. Left: temporal alignment of the knowledge map and video. Right: generation of video, knowledge map, and text modalities via pose estimation, showing kinematic alignment and knowledge-guided synthesis.} 
    \label{fig:1}
\end{figure*}

Technological advances have turned to computer vision for alternative screening. Early methods analyzed static postural asymmetry from back images \cite{Zhang2024New,Zhang2023Deep}, but such approaches cannot capture dynamic kinematic biomarkers that may better reflect AIS biomechanics \cite{Pesenti2020Static}. Subsequent research has utilized gait videos, employing features like silhouette sequences and pose estimation to model movement \cite{Zhou2024Gait,Zhou2025Pose}. The integration of multi-modal learning, successful in other medical domains \cite{King2023Multimodal,Liu2023Medical}, offers a promising direction for AIS. This includes using textual descriptions to guide models and enhance interpretability \cite{Chen2022Multi}. A significant challenge is the high variability in gait presentation. Factors like curve type, compensatory mechanisms, and conscious adjustment by patients result in diverse kinematics \cite{Pizones2024Current,Paramento2024Neurophysiological}. Furthermore, abnormal gaits from other musculoskeletal conditions in non-scoliotic individuals can lead to false positives if models rely on oversimplified patterns \cite{Ji2024Kinematic,Boulcourt2023Gait}. Therefore, effective analysis must account for this heterogeneity to ensure robustness and generalizability. 

Despite promising advances in video-based AIS screening, current approaches face three core limitations: data quality, model interpretability, and integration complexity. First, many benchmark datasets are compromised by data leakage, where multiple video clips from the same subject appear in both training and test sets, allowing models to memorize subject-specific traits rather than learn generalizable pathological patterns. Compounding this issue, training labels are often derived from traditional physical screening methods, which are prone to inaccuracy due to confounding physical factors. Second, deep learning models for complex movements like gait often function as "black boxes," learning features that lack clinical transparency. This limits their interpretability and hinders adoption by practitioners. Finally, while multi-modal fusion with clinical data holds potential, current methods typically employ simplistic fusion mechanisms. These fail to capture the nuanced, clinically relevant relationships between kinematic patterns and auxiliary information, ultimately restricting practical utility.

To address the aforementioned limitations, this work presents three key contributions: 

(1) We introduce ScoliGait, a benchmark dataset designed to eliminate data leakage for reliable evaluation in video-based AIS screening. It comprises 1,572 training clips derived from 550 adolescents and a held-out test set of 300 clips, each from a unique individual unseen during training, ensuring assessment of model generalization. Each clip is directly annotated with its radiographic Cobb angle.

(2) We design a clinical-prior-guided kinematic knowledge map to encode holistic gait dynamics. This structured representation integrates clinical biomarkers of scoliotic kinematics into the learning process, providing an interpretable framework that moves beyond black-box features toward clinician-understandable motion analysis.

(3) We introduce a latent attention pooling mechanism to derive highly expressive, holistic embeddings from kinematic sequences. This technique, inspired by advanced language model embedding methods, utilizes a learnable latent dictionary to query the input sequence via cross-attention. It generates a more powerful and clinically salient representation than standard pooling methods, which serves as a superior foundation for subsequent multimodal fusion with clinical data. 

\section{Related Works}

Recent advances in computer-aided screening for Adolescent Idiopathic Scoliosis have increasingly leveraged video-based analysis. The foundational work established gait as a viable digital biomarker by introducing the Scoliosis1K dataset and ScoNet-MT model, which classifies scoliosis from silhouette sequences of gait videos \cite{Zhou2024Gait}. Building on this, subsequent research sought to incorporate explicit clinical knowledge to improve model interpretability and performance \cite{Zhou2025Pose}. However, such clinical knowledge is not enough to represent holistic gait motion, which is crucial for scoliotic gait analysis and interpretability. The multi-modal learning work by combining visual data with descriptive clinical text frames screening based on kinematic characteristics and clinical prior knowledge \cite{Li2025TextGuidedMIL}. However, the training pipeline is directly fusing representations without sufficient alignments. 

\section{Methods}
\subsection{Dataset Preparation}

The ScoliGait dataset was constructed by aligning with the protocols of the Scoliosis1K benchmark \cite{Zhou2024Gait}. The study enrolled 850 participants who provided informed consent, with diagnostic categorization based on the clinical standard of a 10‑degree Cobb angle. Key demographic characteristics are summarized in Table \ref{tab:demographics}. To expand the training set, each of the 550 original participant videos was segmented into multiple non‑overlapping clips, each capturing several complete walking cycles (96 frames at 30Hz, 1080p resolution). 

\begin{table}[ht!]
\centering
\small
\setlength{\tabcolsep}{3.5pt} 
\renewcommand{\arraystretch}{1.2} 
\caption{Demographic Statistics of the ScoliGait Dataset}
\label{tab:demographics}
\begin{tabular}{|l|c|c|c|c|}
\hline
\multirow{2}{*}{\textbf{Attributes}} & \multicolumn{3}{c|}{\textbf{Diagnostic Category}} \\
\cline{2-4}
& \textbf{All} & \textbf{Positive} & \textbf{Negative} \\
\hline
\makecell{Number of Participants (total)} & 850 & 488 & 362 \\
\hline
\makecell{Gender (Female/Male)} & 456/394 & 384/104 & 72/290 \\
\hline
\makecell{Age (mean $\pm$ std)} & 12.31 $\pm$ 2.87 & 13.29 $\pm$ 2.45 & 11.78 $\pm$ 2.90 \\
\hline
\makecell{Number of Test Set} & 300 & 212 & 88 \\
\hline
\end{tabular}
\end{table}

The kinematic knowledge map (Fig.\ref{fig:1}(Right Middle Row)) was constructed by computing features in motion space, self-skeleton space, and signal cross-correlation, following the clinical gait analysis framework established in prior scoliosis studies \cite{Pizones2024Current,Paramento2024Neurophysiological,Ji2024Kinematic,Boulcourt2023Gait}. This map was subsequently used to temporally align and filter motion clips (Fig.\ref{fig:1}(Right Top Row)), ensuring consistent data curation. The knowledge map and video clips share a direct temporal correspondence, which was explicitly encoded through positional embeddings and utilized for model interpretation (Fig.\ref{fig:1}(Left)). Clinically meaningful textual descriptions were automatically generated to characterize distinct scoliotic gait patterns based on group-level statistical analysis, capturing overarching pathological features such as "asymmetric arm swing in thoracic curve types". These text prompts abstract discriminative kinematic patterns and corresponding effect sizes, grounded in established clinical evidence \cite{Pizones2024Current,Paramento2024Neurophysiological,Ji2024Kinematic,Boulcourt2023Gait}.

The final dataset comprises 1,572 training clips and 300 independent external test clips, with no participant overlap between splits to mitigate the risk of model overfitting to subject‑specific traits. To the best of our knowledge, ScoliGait is the first gait-based scoliosis dataset with labels rigorously validated against radiographic Cobb angle measurements, the clinical gold standard. As shown in Fig.\ref{fig:1}(Right), the dataset includes three modalities: knowledge map, video, and text. All annotations were further verified by senior orthopaedic specialists to ensure diagnostic reliability.

\subsection{Kinematic Knowledge Map}

Previous studies have demonstrated detectable deviations in scoliotic gait from normal patterns \cite{Pizones2024Current,Paramento2024Neurophysiological,Ji2024Kinematic,Boulcourt2023Gait}. To characterize these deviations, we constructed a kinematic knowledge map comprising 238 features across three domains: motion space (140 features), self-skeleton space (32 features), and signal cross-correlation (66 features), as illustrated in Fig.\ref{fig:1}(Right). Paired joint distances were computed using Euclidean metrics, inter‑segmental angles were derived trigonometrically, and temporal coordination, such as limb synchronization, was quantified via signal cross‑correlation.

During inference, the model’s attention scores are mapped back onto this knowledge map, yielding transparent, clinically interpretable explanations. This mapping allows clinicians to intuitively identify which kinematic features the model prioritizes in its assessment. By linking learned attention patterns to established biomechanical concepts, the knowledge map serves as a human-readable dictionary that translates model decisions into actionable insights, bridging algorithmic reasoning with clinical judgment.

\subsection{Model Architecture and Latent Attention Fusion}

Our model architecture comprises three modality-specific encoders for knowledge map, video, and text processing. The text encoder utilizes the Sentence-Transformers framework with the all-MiniLM-L6-v2model \cite{reimers2019sentence,wang2020minilm,sentence-transformers},  while both knowledge map and video encoders adopt a Vision Transformer (ViT) backbone \cite{dosovitskiy2020image}, distinguished only by their patch embedding methods. To maintain comparable parameter scales, we employ 8 transformer layers in single-modality experiments and 4 layers in multimodal settings. Furthermore, inspired by NV-Embed \cite{lee2024nv}, we replace the standard average pooling with a latent attention pooling layer to effectively compress multimodal features during native transformer training.

\begin{figure*}[ht!]
    \includegraphics[width=0.8\textwidth]{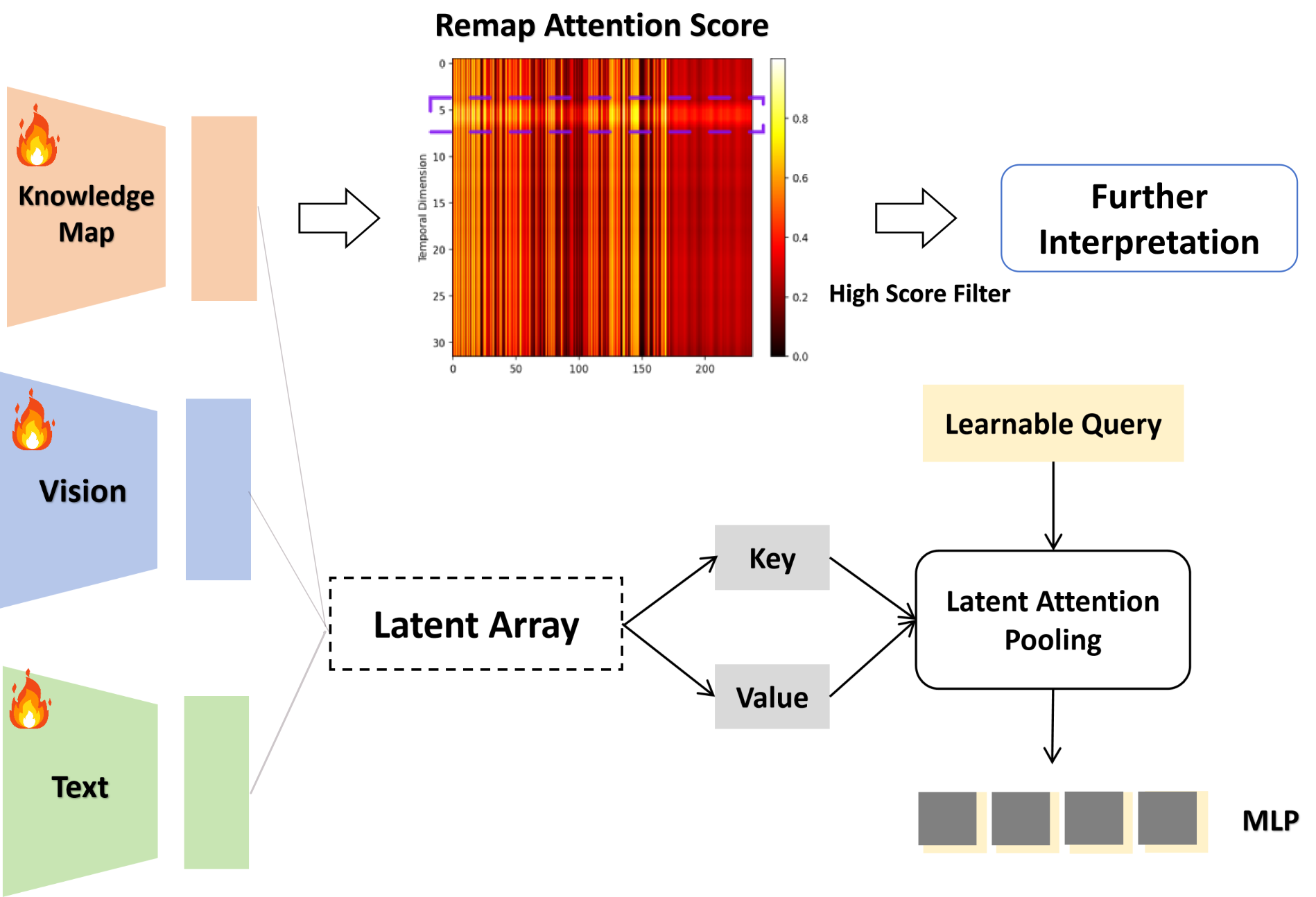}
    \caption{Proposed three-modal fusion architecture for AIS screening. Inputs from Knowledge Map, Vision, and Text modalities are integrated via a Latent Attention Pooling mechanism (bottom). Remapped attention scores from the Knowledge Map (top) are filtered for salient features to enable clinical interpretation. } 
    \label{fig:2}
\end{figure*}

\section{Results}

\subsection{Scoliosis Screening Task}

The performance of different model configurations on the binary AIS screening task is summarized in Table \ref{tab:1}. In single-modality evaluations, the knowledge map model outperformed the video-based model, achieving 1.7\% higher accuracy and a 3.2\% higher F1-score. This result demonstrates that the structured kinematic representation provides a more discriminative encoding of holistic gait motion than raw video features for this task. 

Subsequently, multimodal fusion was evaluated. Integrating the knowledge map with video features yielded a significant performance improvement. Our best-performing model integrates knowledge map, video, and clinical text via the proposed latent attention pooling mechanism, attained the highest performance with accuracy of 70.0\% and F1-score of 61.9\%. 

We also evaluated ScoNet-MT models on our external test set. While their overall accuracy approached that of our two-modality fusion, their metrics for the positive class remained unsatisfactory, indicating a poor capability to correctly identify scoliosis cases.

\begin{table}[ht!]
\centering
\small
\setlength{\tabcolsep}{3.5pt} 
\renewcommand{\arraystretch}{1.2} 
\caption{Binary Classification Performance Comparison of Multiple Methods}
\label{tab:1}
\begin{tabular}{|l|c|c|c|c|c|c|c|c|}
\hline
\multirow{2}{*}{\textbf{Modality}} & \multicolumn{2}{c|}{\textbf{Overall}} & \multicolumn{3}{c|}{\textbf{Positive Class}} & \multicolumn{3}{c|}{\textbf{Negative Class}} \\
\cline{2-9}
& \textbf{Acc} & \textbf{F1} & \textbf{Pre} & \textbf{Rec} & \textbf{F1} & \textbf{Pre} & \textbf{Rec} & \textbf{F1} \\
\hline
\makecell{Video} & 0.563 & 0.488 & 0.278 & 0.306 & 0.291 & 0.699 & 0.669 & 0.684 \\
\hline
\makecell{Knowledge Map} & 0.580 & 0.520 & 0.320 & 0.386 & 0.350 & 0.721 & 0.660 & 0.689 \\
\hline
\makecell{Knowledge Map + \\ Video} & 0.640 & 0.562 & 0.383 & 0.375 & 0.379 & 0.743 & 0.750 & 0.746 \\
\hline
\textbf{\makecell{Knowledge Map + \\ Video + Text}} & \textbf{0.700} & \textbf{0.619} & \textbf{0.486} & \textbf{0.409} & \textbf{0.444} & \textbf{0.769} & \textbf{0.820} & \textbf{0.794} \\
\hline
\hline
\multicolumn{9}{|c|}{\textbf{ScoNet-MT}} \\ 
\hline
\makecell{1:1:2} & 0.644 & 0.439 & 0.333 & 0.054 & 0.093 & 0.694 & 0.900 & 0.784 \\
\hline
\makecell{1:1:4} & 0.648 & 0.457 & 0.275 & 0.087 & 0.132 & 0.697 & 0.891 & 0.782 \\
\hline
\makecell{1:1:8} & 0.651 & 0.416 & 0.166 & 0.021 & 0.038 & 0.695 & 0.924 & 0.793 \\
\hline
\makecell{1:1:16} & 0.664 & 0.457 & 0.318 & 0.076 & 0.122 & 0.696 & 0.919 & 0.792  \\
\hline
\end{tabular}
\end{table}

\subsection{Explainability}

This paradigm offers significant advantages by integrating clinically interpretable knowledge directly into the model architecture. The kinematic knowledge map provides a sparse, structured representation of gait features, delivering both higher temporal resolution and an explicitly interpretable clinical format. Unlike common post-hoc methods that simply highlight areas of attention on raw video frames, this map functions as a queryable dictionary for clinicians. It elucidates the specific kinematic patterns, such as asymmetry or specific joint dynamics, offering reasoning beyond mere localization. These model-identified features can then be validated through established physical examination techniques.

\begin{figure*}[h]
    \includegraphics[width=\textwidth]{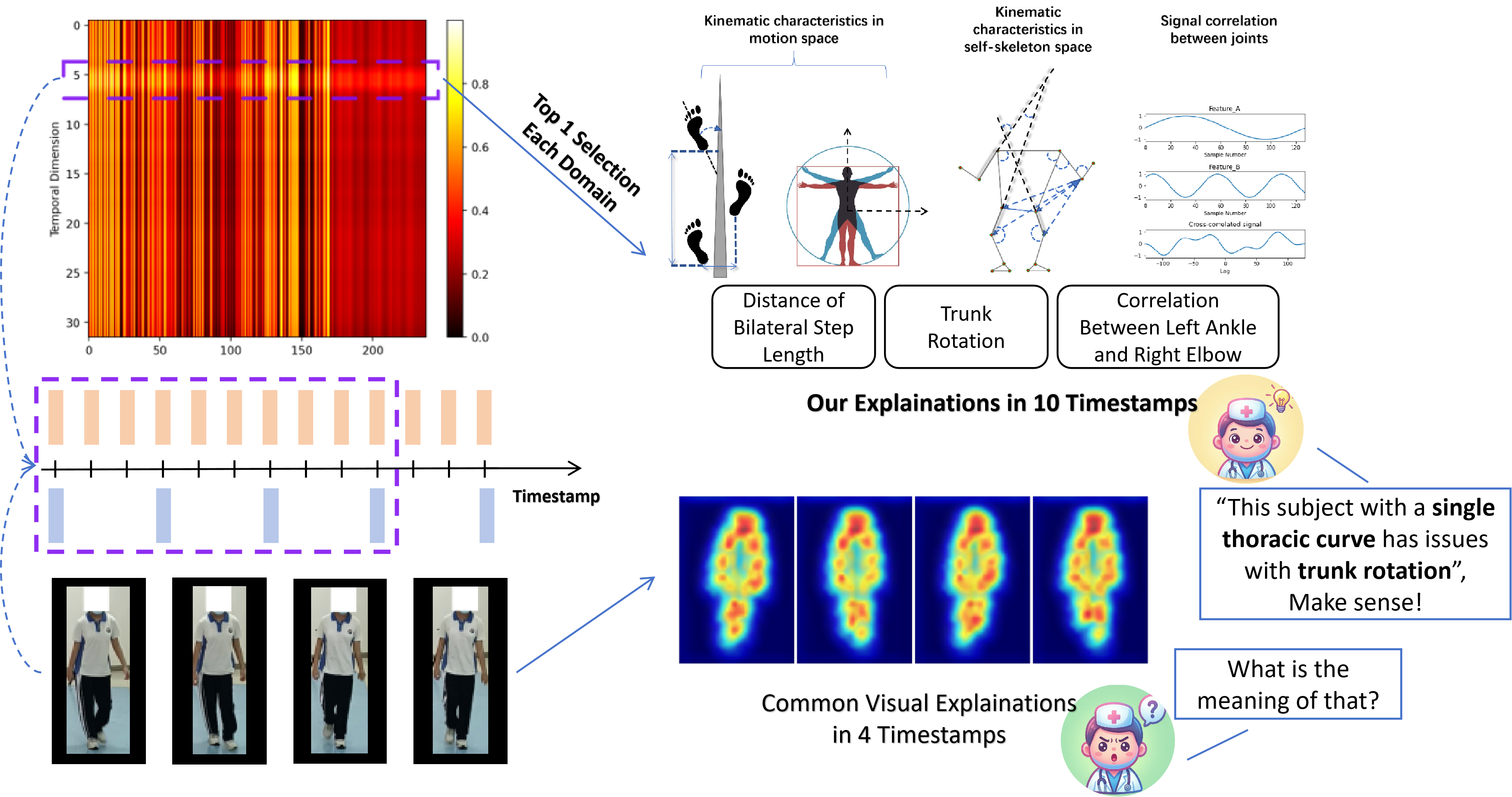}
    \caption{Comparison of interpretability methods. Left: The proposed kinematic knowledge map (top) with aligned video (bottom), where axes are time vs. clinical variables. Right: Explicit kinematic features extracted from the map (top) versus conventional pose-based attention heatmaps (bottom). Our method provides explicit clinical insights beyond spatial saliency.} 
    \label{fig:3}
\end{figure*}

For instance, as illustrated in Fig. \ref{fig:3}, the dashed rectangle highlights a region of high model activation within the knowledge map, where the axes represent time and clinical kinematic variables, respectively. This enables more granular temporal analysis than the source video frames alone. By selecting the top-ranked feature in each relevant domain, where the column index corresponds to a specific clinical prior feature and the row index indicates its time period, the knowledge map delivers explicit, high-resolution interpretations. In contrast, conventional attention maps overlaid on human skeletons often fail to articulate the underlying clinical rationale for the model’s focus.

\subsection{Ablation Studies}

We conducted ablation studies on two key design choices: the multimodal fusion strategy and the alignment of positional embeddings between the knowledge map and video modalities (Table \ref{tab:2}). Our evaluation of fusion methods shows the proposed latent attention pooling (Cat+Latent) outperforms both simple concatenation (Cat) and concatenation with a standard attention layer (Cat+Att). The Cat+Latent configuration achieved the best overall performance (Accuracy: 64\%, F1: 56.2\%). Furthermore, aligning positional embeddings across modalities improved model performance. Applying the same Cat+Latent fusion with non-aligned embeddings led to a measurable decline in key metrics, confirming the utility of our alignment strategy.

\begin{table}[ht!]
\centering
\small
\setlength{\tabcolsep}{2.5pt}
\renewcommand{\arraystretch}{1.4}
\caption{Binary classification performance of different multimodal fusion strategies and positional embedding alignments between the knowledge map and video modalities.}
\label{tab:2}
\begin{tabular}{|l|l|c|c|c|c|c|c|c|c|}
\hline
\multirow{2}{*}{\textbf{RoPE}} & \multirow{2}{*}{\textbf{Config}} & \multicolumn{2}{c|}{\textbf{Overall}} & \multicolumn{3}{c|}{\textbf{Positive Class}} & \multicolumn{3}{c|}{\textbf{Negative Class}} \\
\cline{3-10}
& & \textbf{Acc} & \textbf{F1} & \textbf{Pre} & \textbf{Rec} & \textbf{F1} & \textbf{Pre} & \textbf{Rec} & \textbf{F1} \\
\hline
\multirow{3}{*}{\textbf{Aligned}} & \makecell{Concat(Cat)} & 0.550 & 0.495 & 0.292 & 0.375 & 0.328 & 0.705 & 0.622 & 0.661 \\
\cline{2-10}
& \makecell{Cat+Att} & 0.610 & 0.531 & 0.337 & 0.341 & 0.339 & 0.725 & 0.721 & 0.723 \\
\cline{2-10}
& \makecell{\textbf{Cat+Latent}} & \textbf{0.640} & \textbf{0.562} & \textbf{0.383} & \textbf{0.375} & \textbf{0.379} & \textbf{0.743} & \textbf{0.750} & \textbf{0.746} \\
\hline
\makecell{\textbf{Non-}\\ \textbf{Aligned}} & \makecell{Cat+Latent} & 0.623 & 0.531 & 0.341 & 0.306 & 0.323 & 0.724 & 0.754 & 0.739 \\
\hline
\end{tabular}
\end{table}

\section{Conclusions}

This paper presents a comprehensive framework for video-based adolescent idiopathic scoliosis (AIS) screening, systematically addressing key limitations in data reliability, model interpretability, and multimodal integration. We introduced ScoliGait, a novel benchmark dataset curated to eliminate data leakage by ensuring unique individuals in the test set, paired with radiographic Cobb angle labels and clinical text descriptions. To move beyond "black-box" models, we proposed a clinical-prior-guided kinematic knowledge map, which provides an interpretable, sparse representation of gait dynamics directly aligned with clinical biomarkers. Furthermore, we introduced a latent attention pooling mechanism for fusion of video, knowledge map, and textual modalities, enabling the model to learn nuanced, clinically-relevant representations.

\clearpage
\bibliographystyle{splncs04}
\bibliography{reference}

\end{document}